\documentclass{article}

\usepackage[final]{neurips_2024}

\usepackage[utf8]{inputenc} 
\usepackage[T1]{fontenc}    
\usepackage{hyperref}       
\usepackage{url}            
\usepackage{booktabs}   
\usepackage{graphicx}
\usepackage{amsfonts}  
\usepackage{algorithm}
\usepackage{amsmath}
\usepackage{algpseudocode}
\usepackage{nicefrac}       
\usepackage{microtype}      
\usepackage{xcolor}         
\usepackage{cite}
\usepackage{float}
\usepackage[most]{tcolorbox}

\title{MIRROR: A Novel Approach for the Automated Evaluation of Open-Ended Question Generation}

\author{%
   Aniket Deroy\\
   Computer Science and Engineering\\
   IIT Kharagpur\\
  \texttt{roydanik18@kgpian.iitkgp.ac.in} \\
   \And
   Subhankar Maity \\
   Department of Artificial Intelligence\\
   IIT Kharagpur \\
   \texttt{subhankar.ai@kgpian.iitkgp.ac.in} \\
   \And
   Sudeshna Sarkar\\
   Computer Science and Engineering\\
   IIT Kharagpur\\
  \texttt{sudeshna@cse.iitkgp.ac.in} \\
}

\begin{document}

\maketitle

\begin{abstract}
Automatic question generation is a critical task that involves evaluating question quality by considering factors such as engagement, pedagogical value, and the ability to stimulate critical thinking. These aspects require human-like understanding and judgment, which automated systems currently lack. However, human evaluations are costly and impractical for large-scale samples of generated questions. Therefore, we propose a novel system, MIRROR (Multi-LLM Iterative Review and Response for Optimized Rating), which leverages large language models (LLMs) to automate the evaluation process for questions generated by automated question generation systems. We experimented with several state-of-the-art LLMs, such as GPT-4, Gemini, and Llama2-70b. We observed that the scores of human evaluation metrics, namely relevance, appropriateness, novelty, complexity, and grammaticality, improved when using the feedback-based approach called MIRROR, tending to be closer to the human baseline scores. Furthermore, we observed that Pearson's correlation coefficient between GPT-4 and human experts improved when using our proposed feedback-based approach, MIRROR, compared to direct prompting for evaluation. Error analysis shows that our proposed approach, MIRROR, significantly helps to improve relevance and appropriateness.

\end{abstract}

\section{Introduction}

Automated question generation (AQG) is crucial in education because it enhances critical thinking, promotes active learning, and provides personalized learning experiences. Currently, metrics such as BLEU, METEOR, and ROUGE are used for the purpose of evaluation of open ended questions \citep{r6, r7}.
However, traditional automated evaluation metrics such as BLEU, ROUGE, or METEOR, which are often used for tasks like machine translation or summarization, may not be well-suited for evaluating the quality of generated questions \citep{r2, r3, r4}. These metrics primarily measure surface-level similarity to reference questions, rather than deeper aspects of question quality such as relevance, clarity, and engagement \citep{r1}. Some aspects of question quality, such as engagement, pedagogical value, and the potential to stimulate critical thinking, require a human-like understanding and judgment that automated systems cannot yet replicate. However human judgements are costly and cannot be replicated over large samples of generated questions. So we propose an LLM feedback-based system called \textbf{MIRROR} 
 (\textit{Multi-LLM Iterative Review and Response for Optimized Rating}) for automating the process of human evaluation for questions generated by AQG systems. We understand that LLMs have understanding and reasoning capabilities. 
 
 So, the question we ask is \textit{Can LLMs replace human experts in evaluating the quality of generated questions by AQG systems?} The results show that our proposed feedback-based approach for evaluating questions is worthwhile in generating quality evaluation for questions generated by AQG systems.
We observe that the scores of the human evaluation metrics namely \textit{relevance} \citep{r5}, \textit{appropriateness} \citep{r5}, \textit{novelty} \citep{r5}, \textit{complexity} \citep{r14, r5}, and \textit{grammaticality} \citep{r15, r5, r16} improve on using the feedback-based approach called MIRROR and tend to be closer to the human baseline scores. Also, we observe that the Pearson's correlation coefficient between GPT-4 and human experts improves on using our proposed feedback-based approach called MIRROR as compared to using direct prompting for evaluation.

Our contributions in this work are as follows:

\begin{enumerate}

\item We propose a novel method, \textbf{MIRROR} (\textbf{\underline{M}}ulti-LLM \textbf{\underline{I}}terative \textbf{\underline{R}}eview and \textbf{\underline{R}}esponse for \textbf{\underline{O}}ptimized \textbf{\underline{R}}ating). This feedback-based procedure involves prompting various LLMs to generate scores based on human evaluation metrics, namely \textit{grammaticality}, \textit{relevance}, \textit{appropriateness}, \textit{novelty}, and \textit{complexity}, for the task of automated open-ended question generation. We then ask the same model to list the strengths and flaws of the provided questions. These strengths and flaws are subsequently given to another LLM along with the questions, prompting it to repeat the evaluation procedure.

\item  We show that our proposed LLM feedback-based approach, which provides scores based on various human evaluation metrics, produces results closer to the human baseline compared to the direct prompting approach.

\item We also measure the Pearson’s correlation coefficient between the GPT-4 generated scores and those of human evaluators, demonstrating that the feedback-based approach improves correlation compared to the direct approach for evaluating generated questions. Error analysis shows that our proposed approach, MIRROR, significantly improves the \textit{relevance} and \textit{appropriateness} of the evaluations.

\end{enumerate}

\section{Related Work}

Evaluating open-ended question generation using automated metrics (e.g., BLEU, ROUGE, METEOR, etc.) presents significant challenges \citep{r1, r14, r15, r16}. These metrics often fail to capture higher-order cognitive skills and overlook deeper educational values such as stimulating critical thinking \citep{r7, r9}. Automated evaluations may not fully account for nuances such as context relevance and cognitive complexity, making it difficult to assess whether questions promote skills such as analysis, synthesis, and evaluation \citep{r9}. Furthermore, scalability issues arise because human evaluation for large datasets is impractical \citep{r7}. Using LLMs to evaluate human-like criteria such as \textit{grammaticality}, \textit{relevance}, \textit{appropriateness}, \textit{novelty}, and \textit{complexity} has been a prominent research area \citep{r12}. Aligning machine-generated content with human judgment is crucial for practical applicability. Incorporating feedback loops within LLMs to refine their outputs is an emerging field \citep{r11, r17, r18, r19}. Techniques like Reinforcement Learning from Human Feedback (RLHF) fine-tune LLMs based on human preferences, enhancing the quality and contextual relevance of responses \citep{r10, r17, r20}. This work proposes a novel method using LLM feedback to evaluate open-ended questions based on human-like metrics: \textit{grammaticality} \citep{r15, r5, r16}, \textit{relevance} \citep{r5}, \textit{appropriateness} \citep{r5}, \textit{complexity} \citep{r14, r5}, and \textit{novelty} \citep{r5}. This approach aims to bridge the gap between automated metrics and human evaluation, enhancing the quality and applicability of generated questions in educational settings.

\section{Dataset}
We use 1000 samples from the EduProbe dataset \citep{r5} for our experiments. The dataset consists of <\textit{Context}, \textit{Question}> pairs from subjects such as History, Geography, Economics, Environmental Studies, and Science. For EduProbe, we already have the gold-standard questions corresponding to the context. We also use 500 samples from the SciQ dataset \citep{r8} for our experiments. The dataset consists of <\textit{Context}, \textit{Question}> pairs from subjects such as Physics, Chemistry, Biology, and Earth Science. For SciQ dataset we create a question corresponding to the context by using an educator.
The contexts from both these datasets are considered for our experiments to generate the questions using the LLMs and then to evaluate them automatically via LLMs.
Together EduProbe and SciQ datasets covers a wide domain of subjects necessary for generating open-ended questions and showing the wider applicability of our work.


\section{Methodology}
In this section, we discuss the generation of questions via prompting GPT-3.5 Turbo (Section \ref{one}), the direct approach to evaluating the quality of the generated questions (Section \ref{two}), the feedback-based approach for evaluating generated question quality (Section \ref{three}), and the correlation between the best-performing LLM and human experts (Section \ref{four}). In our experiments, we used state-of-the-art LLMs, such as GPT-4 \citep{r21}, Gemini \citep{r22}, and Llama2-70b \citep{r23}, to assess both direct and feedback-based approaches.


\subsection{Generating Questions via Prompting}\label{one}
\begin{figure*}[h]
  \centering
  \includegraphics[width=\linewidth]{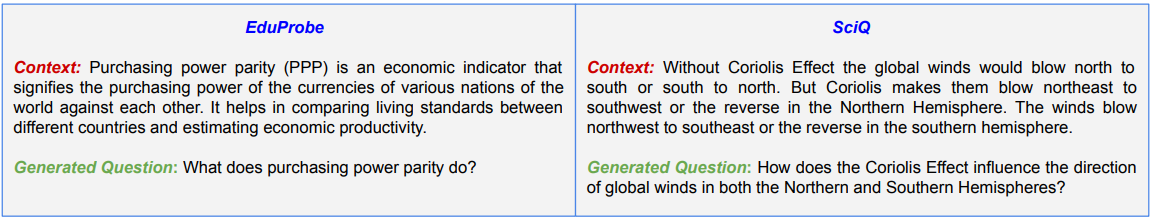}
  \caption{A sample of <\textit{Context}, \textit{Generated Question}> pairs from the EduProbe and SciQ datasets.} 
  \label{s1}
\end{figure*}

\begin{figure*}[h]
  \centering
  \includegraphics[width=0.50\linewidth]{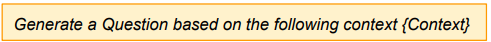}
  \caption{Prompt used on GPT-3.5 Turbo to generate a question from a context.} 
  \label{s2}
\end{figure*}

Figure~\ref{s1} provides <\textit{Context}, \textit{Generated Question}> pair from the EduProbe and SciQ dataset respectively. Figure~\ref{s2} provides the prompt for generating questions from a context. We prompt GPT-3.5 Turbo to generate question from a context corresponding to EduProbe and SciQ datasets. We evaluate the generated questions based on five metrics: \textit{grammaticality}, \textit{appropriateness}, \textit{relevance}, \textit{novelty}, and \textit{complexity}.

\subsection{Direct Approach for Evaluating Question Quality}\label{two}
\begin{figure*}[h!]
  \centering
  \includegraphics[width=0.80\linewidth]{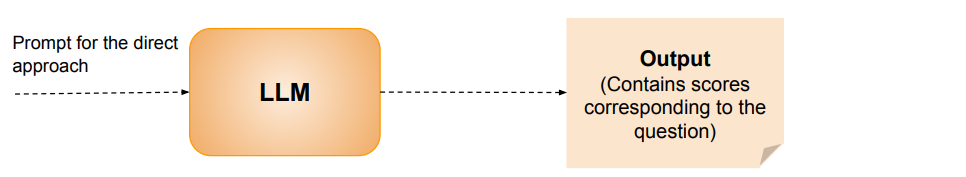}
  \caption{An overview of the direct prompting approach.} 
  \label{s7}
\end{figure*}

\begin{figure*}[h!]
  \centering
  \includegraphics[width=0.90\linewidth]{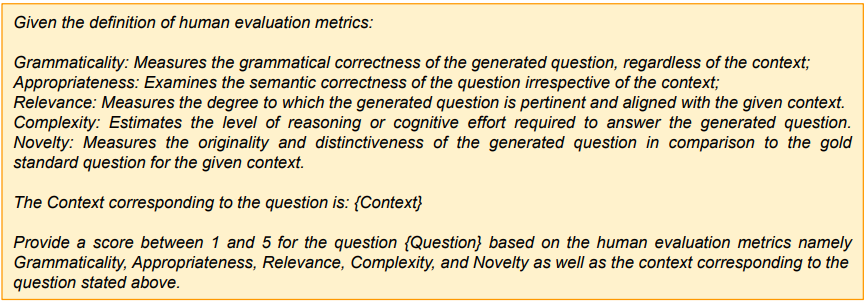}
  \caption{Prompt used in direct approach for evaluating human evaluation metrics.} 
  \label{s8}
\end{figure*}

The algorithm~\ref{s6} provides the set of steps required to produce the direct approach. The overview diagram for our direct approach is shown in Figure~\ref{s7}. The prompt provided to the LLMs for the direct prompting approach is shown in Figure~\ref{s8}.

\begin{algorithm}
\footnotesize	
\caption{Direct Approach}\label{alg:single_llm}
\begin{algorithmic}
\footnotesize	
\Require Human Evaluation Metric definitions
\Require Question and its context
\Ensure Scores

\State Compute scores for human evaluation metrics via LLM prompting provided the given context, question, and metrics.


\end{algorithmic}
\label{s6}
\end{algorithm}

\subsection{Feedback-based Approach for Evaluating Question Quality}\label{three}
The input to our proposed MIRROR approach includes human evaluation metric definitions, the initial strengths set ($S_0$), the initial flaws set ($F_0$), a generated question to be evaluated, and its corresponding context. The output from our approach is the final human evaluation metric scores. The goal of our algorithm is to perform repeated feedback between two LLMs (i.e., $LLM_1$, $LLM_2$) so that the metric scores converge. The strengths and flaws generated by the LLMs are extracted from the entire output using a rule-based pattern-matching algorithm. The algorithm~\ref{s3} provides the set of steps required to produce the MIRROR approach. 

Initially, the process starts by defining a set of human evaluation metrics which are the criteria used to assess the quality of the question. These metrics include \textit{grammaticality}, \textit{appropriateness}, \textit{relevance}, \textit{complexity}, and \textit{novelty} which contribute to the effectiveness of the question.
At the outset, two sets are initialized: one for strengths ($S_0$) and another for flaws ($F_0$). Both sets start empty. The next step involves computing initial scores for the question based on the predefined evaluation metrics. From this assessment by $LLM_1$, the first sets of strengths and flaws are identified, resulting in $S_1$ and $F_1$.
The process then enters an iterative loop designed to refine these initial assessments. In each iteration, the identified strengths and flaws ($S_1$, $F_1$) are provided as feedback to the second LLM, $LLM_2$. Along with this feedback, the human evaluation metrics, the question itself, and its context are also provided. $LLM_2$ then generates new scores for the evaluation metrics and updates the sets of strengths and flaws to $S_2$ and $F_2$.
The updated strengths and flaws from $LLM_2$ (i.e., $S_2$, $F_2$) are then fed back to $LLM_1$. $LLM_1$ uses this information to re-evaluate the question, updating its scores and further refining the sets of strengths and flaws to $S_3$ and $F_3$.
This process continues iteratively, with each model using the feedback from the other to refine its evaluation, until a convergence criterion is met. The convergence is typically defined as the point at which the scores from $LLM_1$ and $LLM_2$ become identical for two consecutive iterations. Once convergence is achieved, the loop terminates, and the final, converged scores and associated strengths and flaws are considered the accurate evaluation of the question. This iterative process ensures that the evaluation is thorough and benefits from the complementary perspectives of two different LLMs, ultimately leading to a more reliable and nuanced assessment of the question.

\begin{figure*}[h!]
  \centering
  \includegraphics[width=0.94\linewidth]{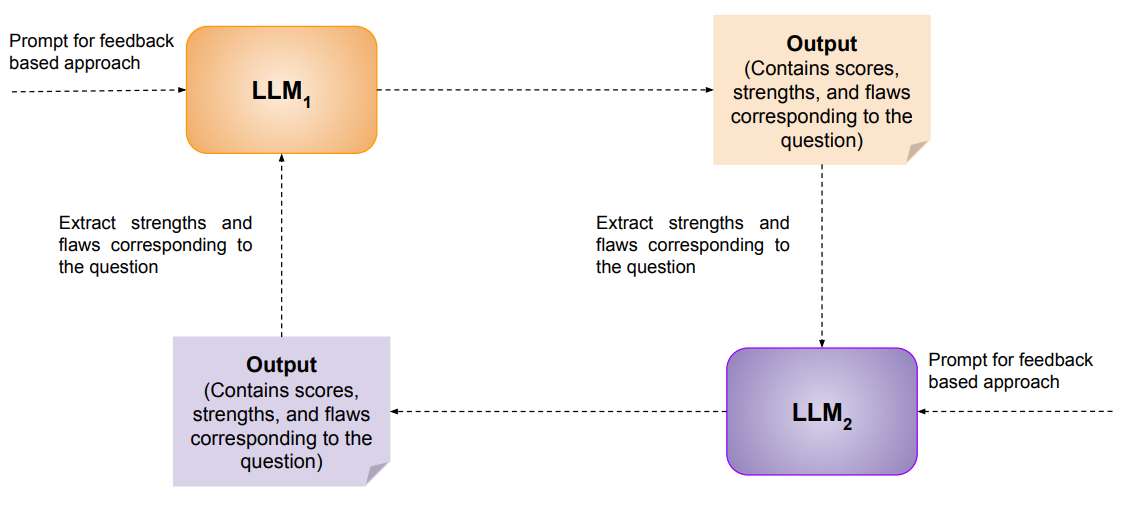}
  \caption{An overview of the proposed approach called MIRROR.} 
  \label{s4}
\end{figure*}

\begin{figure*}[h!]
  \centering
  \includegraphics[width=0.90\linewidth]{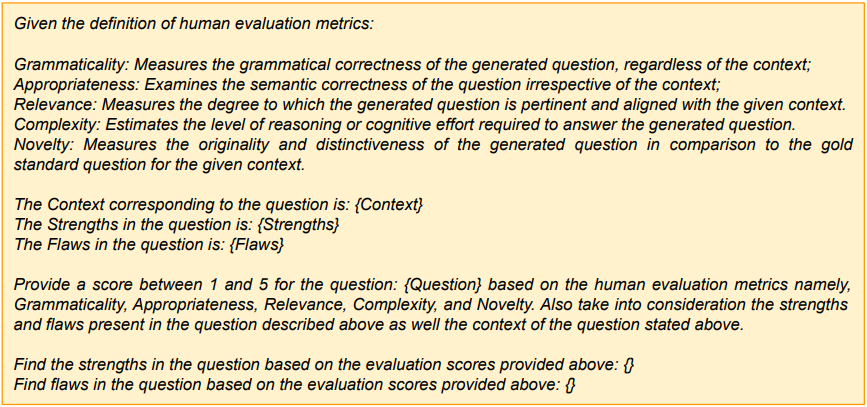}
  \caption{Feedback-based prompt for evaluating human evaluation metrics.} 
  \label{s5}
\end{figure*}

The overview diagram for our MIRROR approach is shown in Figure~\ref{s4}. The prompt provided to the LLMs for producing the MIRROR approach is shown in Figure~\ref{s5}. The method described is a process for refining the evaluation of a question using a feedback loop between two LLMs. The goal is to achieve a consensus on the evaluation scores and identify key strengths and flaws in the question, ensuring that the evaluation is as accurate and comprehensive as possible.







\begin{algorithm}
\footnotesize	
\caption{MIRROR (Multi-LLM Iterative Review and Response for Optimized Rating)}\label{alg:mirror}
\begin{algorithmic}
\footnotesize	
\Require Human Evaluation Metric definitions
\Require Initial strengths set $S_0$ and flaws set $F_0$, both initialized as empty sets
\Require Question and its context
\Ensure Converged scores for the question

\State $S_0 \gets \{\}$
\State $F_0 \gets \{\}$
\State Compute initial scores for human evaluation metrics with the given question and context

\State \textbf{Identify initial strengths and flaws}
\State $S_1, F_1 \gets \text{Identify strengths and flaws}$

\While{convergence criteria are not met}
    \State Provide $S_1$, $F_1$ as feedback, along with metrics, question, and context, to $LLM_2$
    \State Ask $LLM_2$ to generate scores for human evaluation metrics
    
    \State Obtain strengths $S_2$ and flaws $F_2$ from $LLM_2$
    
    \State Provide $S_2$, $F_2$ as feedback, along with metrics, question, and context, to $LLM_1$
    \State Ask $LLM_1$ to provide updated scores for the question
    \State Obtain updated strengths $S_3$ and flaws $F_3$ from $LLM_1$
    \State Update: $S_1 \gets S_3$, $F_1 \gets F_3$\;
    \State \textbf{Check for convergence}
    \If{scores from $LLM_1$ and $LLM_2$ are identical for two consecutive iterations}
        \State \textbf{Terminate the loop}
    \EndIf
\EndWhile

\end{algorithmic}
\label{s3}
\end{algorithm}

\subsection{Human Baseline Scores and Pearson's Correlation Coefficient}\label{four}
We recruit three educators and ask them to evaluate the questions generated by prompting GPT-3.5 Turbo in terms of \textit{grammaticality}, \textit{appropriateness}, \textit{relevance}, \textit{novelty}, and \textit{complexity} to generate the human baseline scores corresponding to EduProbe and SciQ datasets. The three educators were asked to provide scores for each metric on a scale of 1 to 5. The scores given by the educators for each metric were then averaged. The inter-annotator agreement is calculated in terms of Fleiss's Kappa. The inter-annotator agreement between the three experts is 0.67, 0.64, 0.66, 0.45, and 0.59 for \textit{grammaticality}, \textit{appropriateness}, \textit{relevance}, \textit{novelty}, and \textit{complexity}, respectively. Later we also calculate the Pearson's correlation coefficient \citep{r13} between the best-performing model (i.e., GPT-4) and human experts.


\section{Results}

\textbf{Human Evaluation Results and Correlation Analysis.}
Table~\ref{tab:metrics_table1} shows the human evaluation metric scores for the EduProbe dataset. Table~\ref{tab:metrics_t2} shows the human evaluation metric scores for the SciQ dataset. Table~\ref{tab:metrics_t3} shows the Pearson's correlation coefficient scores for the EduProbe and SciQ datasets between GPT-4 and human experts. We observe that the human baseline scores the highest in \textit{grammaticality}, \textit{relevance}, and \textit{appropriateness}, but lower in \textit{novelty} and \textit{complexity} for both the EduProbe and SciQ datasets. GPT-4 performs closest to the human baseline, especially in the feedback-based approach, with high scores in \textit{grammaticality} and \textit{appropriateness}. Gemini consistently scores slightly lower than GPT-4 across all metrics and evaluation methods. Llama2-70b has the lowest scores among the LLMs explored but still performs relatively well, with its highest score in \textit{grammaticality}. Overall, LLMs tend to score higher in the feedback-based approach compared to the direct approach, suggesting that feedback helps improve perceived performance. In summary, GPT-4 outperforms Gemini and Llama2-70b in both evaluation methods, with scores closest to human performance, especially in \textit{grammaticality} and \textit{appropriateness}. Gemini performs moderately well, while Llama2-70b shows relatively lower scores across the metrics. For the EduProbe and SciQ datasets, GPT-4 shows higher correlation scores with human experts when using the feedback-based approach compared to the direct approach across all metrics. The highest correlation score is for \textit{grammaticality} with a feedback-based approach. For the feedback-based approach, \textit{appropriateness} and \textit{relevance} show moderate correlations, suggesting a reasonable alignment with human expert evaluations in these areas. We observe lower correlations in \textit{novelty} and \textit{complexity} because both approaches (i.e., feedback-based and direct) show lower correlations in these areas, with the lowest being \textit{complexity} in the direct approach. The direct approach generally yields lower correlation scores across all metrics, highlighting the importance of feedback in improving GPT-4’s alignment with human expert evaluations. In summary, the correlation scores suggest that GPT-4’s evaluations are more aligned with human experts when using the feedback-based approach, particularly in \textit{grammaticality}. The alignment is moderate in \textit{appropriateness} and \textit{relevance}, while it is weaker in \textit{novelty} and \textit{complexity}. Direct approach results in overall lower correlation scores across all metrics.

For the SciQ and EduProbe datasets, the feedback-based approach achieves a higher correlation with human evaluations in terms of \textit{grammaticality}, \textit{appropriateness}, \textit{relevance}, \textit{novelty}, and \textit{complexity}, suggesting that iteratively refining the questions using LLM feedback better aligns with human judgments on grammatical correctness (See Table~\ref{tab:metrics_t3}). For \textit{appropriateness}, the feedback-based approach shows a modest improvement over the direct approach. This indicates that feedback helps in evaluating whether the questions are suitable and contextually appropriate more closely to how humans would rate them. The feedback-based approach significantly outperforms direct prompting in \textit{relevance}, suggesting that iterative LLM feedback effectively enhances the model’s ability to judge how relevant the questions are to the given context. The feedback-based method also achieves a higher correlation in assessing \textit{novelty}. This indicates that feedback helps the model better understand and evaluate the uniqueness and originality of the questions. While the improvement in \textit{complexity} evaluation is smaller, the feedback-based approach still surpasses direct prompting. This suggests that feedback helps the model slightly better assess the intricacy and difficulty of the questions. Overall, the higher correlation scores for the feedback-based approach across all metrics suggest that iteratively using LLM feedback improves the alignment of GPT-4’s evaluations with human expert judgments, making it a more effective method for assessing the quality of generated questions.

\begin{table}
\scriptsize
\centering
\caption{Human evaluation results on the EduProbe dataset for grammaticality (Gram), appropriateness (App), relevance (Rel), novelty (Nov), and complexity (Com). \textcolor{blue}{Blue} indicates the highest metric values for the corresponding methods.}
\begin{tabular}{|l|lllll|}
\hline
\textbf{Model} &  \textbf{Gram} & \textbf{App}  & \textbf{Rel} & \textbf{Nov} & \textbf{Com} 
\\
\hline






Human Baseline & 4.95 & 4.97 & 4.48 & 3.98 & 3.10 \\ \hline



\multicolumn{6}{|c|}{\textit{Eduprobe (Direct Approach)}} \\ \hline

GPT-4  & \textcolor{blue}{4.81} & \textcolor{blue}{4.73} & \textcolor{blue}{4.20} & \textcolor{blue}{4.12} & \textcolor{blue}{3.92} \\ \hline

Gemini  & 4.61 & 4.51 & 4.02 & 4.03 & 3.88  \\ \hline

Llama2-70b & 4.38 & 4.20 & 3.84 & 4.01 & 3.88  \\ \hline

\multicolumn{6}{|c|}{\textit{EduProbe (Feedback-based Approach) }} \\ \hline

GPT-4  & \textcolor{blue}{4.87} & \textcolor{blue}{4.82} & \textcolor{blue}{4.30} & \textcolor{blue}{4.25} & \textcolor{blue}{4.05} \\ \hline

Gemini  & 4.72 & 4.64 & 4.14 & 4.10 & 4.00  \\ \hline

Llama2-70b & 4.60 & 4.62 & 4.08 & 4.06 & 3.83  \\ \hline




















\end{tabular}


\label{tab:metrics_table1}
\end{table}

\begin{table}
\scriptsize
\centering
\caption{Human evaluation results on the SciQ dataset for grammaticality (Gram), appropriateness (App), relevance (Rel), novelty (Nov), and complexity (Com). \textcolor{blue}{Blue} indicates the highest metric values for the corresponding methods.}
\begin{tabular}{|l|lllll|}
\hline
\textbf{Model} &  \textbf{Gram} & \textbf{App}  & \textbf{Rel} & \textbf{Nov} & \textbf{Com} 
\\
\hline






Human Baseline & 4.90 & 4.93 & 4.38 & 3.99 & 3.20 \\ \hline



\multicolumn{6}{|c|}{\textit{SciQ (Direct Approach)}} \\ \hline

GPT-4  & \textcolor{blue}{4.70} & \textcolor{blue}{4.44} & \textcolor{blue}{4.03} & \textcolor{blue}{4.01} & \textcolor{blue}{3.74} \\ \hline

Gemini  & 4.42 & 4.34 & 3.92 & 3.84 & 3.65  \\ \hline

Llama2-70b & 4.23 & 4.10 & 3.73 & 3.67 & 3.28  \\ \hline

\multicolumn{6}{|c|}{\textit{SciQ (Feedback-based Approach) }} \\ \hline

GPT-4  & \textcolor{blue}{4.77} & \textcolor{blue}{4.74} & \textcolor{blue}{4.24} & \textcolor{blue}{4.20} & \textcolor{blue}{4.01} \\ \hline

Gemini  & 4.64 & 4.58 & 4.08 & 4.04 & 3.93  \\ \hline

Llama2-70b & 4.58 & 4.55 & 3.94 & 3.91 & 3.80  \\ \hline




















\end{tabular}


\label{tab:metrics_t2}
\end{table}

\begin{table}
\scriptsize
\centering

\caption{Pearson’s correlation coefficient scores for the EduProbe and SciQ datasets between GPT-4 and human experts on grammaticality (Gram), appropriateness (App), relevance (Rel), novelty (Nov), and complexity (Com). \textcolor{blue}{Blue} denotes the highest correlation values for a particular dataset and approach.}

\begin{tabular}{|l|lllll|}
\hline
\textbf{Model} &  \textbf{Gram} & \textbf{App}  & \textbf{Rel} & \textbf{Nov} & \textbf{Com} 
\\
\hline





\multicolumn{6}{|c|}{\textit{EduProbe}} \\ \hline

GPT-4 (Direct Approach) & 0.41 & 0.32 & 0.26 & 0.25 & 0.28 \\ \hline

GPT-4 (Feedback-based Approach) & \textcolor{blue}{0.62} & \textcolor{blue}{0.48} & \textcolor{blue}{0.51} & \textcolor{blue}{0.42} & \textcolor{blue}{0.38} \\ \hline

\multicolumn{6}{|c|}{\textit{SciQ}} \\ \hline

GPT-4 (Direct Approach) & 0.40 & 0.36 & 0.30 & 0.32 & 0.30 \\ \hline

GPT-4 (Feedback-based Approach) & \textcolor{blue}{0.65} & \textcolor{blue}{0.44} & \textcolor{blue}{0.55} & \textcolor{blue}{0.44} & \textcolor{blue}{0.34} \\ \hline




















\end{tabular}


\label{tab:metrics_t3}
\end{table}

\textbf{Error Analysis.}
We conducted a human study using questions from the EduProbe and SciQ datasets, which were evaluated by each LLM using both feedback-based and direct approaches. We observed that the scores provided by human experts and the direct approach matched in 54\%, 39\%, 36\%, 42\%, and 46\% of the cases for \textit{grammaticality}, \textit{relevance}, \textit{appropriateness}, \textit{complexity}, and \textit{novelty}, respectively. We observed that the scores provided by human experts and the feedback-based approach matched in 67\%, 64\%, 62\%, 55\%, and 61\% for \textit{grammaticality}, \textit{relevance}, \textit{appropriateness}, \textit{complexity}, and \textit{novelty}, respectively. This suggests that the direct approach has significant differences compared to human experts, and the use of the feedback-based approach significantly improves the results, especially in terms of \textit{relevance} and \textit{appropriateness}. So, \textit{relevance} and \textit{appropriateness} are two areas where our feedback-based approach (i.e., MIRROR) significantly improves, in addition to moderate improvements in other metrics.

\section{Examples of Outputs Generated by LLMs Using the Feedback-based Approach}
The feedback approach consists of feedback occurring between two LLMs, namely $LLM_1$ and $LLM_2$.
Figure ~\ref{tab:example29} shows the output provided by $LLM_1$ on a generated question based on Economics framed from the EduProbe dataset. Figure ~\ref{tab:example30} shows the output provided by $LLM_2$ on a generated question based on Economics framed from the EduProbe dataset. We observe that the scores provided by $LLM_1$ and $LLM_2$ for different human evaluation metrics have become the same after completion of the Feedback Approach.

The context involved in the question generation process is: \textit{"Purchasing power parity (PPP) is an economic indicator that signifies the purchasing power of the currencies of various nations of the world against each other. It helps in comparing living standards between different countries and estimating economic productivity."}. The question generated by GPT-3.5 Turbo is: \textit{"What does purchasing power parity do?"}
\begin{figure}[tb]
    \centering
     \begin{tcolorbox}
     [enhanced, fit to height=16cm, 
     colback=teal!5!white, colframe=teal!!white, drop fuzzy shadow]

      {\fontfamily{qcr}\selectfont

      Scores and Rationale for the Question: "What does purchasing power parity do?" \\ 

    Grammaticality: 5/5
    
        Strengths: The question is well-formed and free from grammatical errors.

        Flaws: None.\\

    Appropriateness: 4.5/5
    
        Strengths: The question is semantically correct and understandable.
        
        Flaws: The question is very general and could be more specific about PPP.\\

    Relevance: 5/5
    
        Strengths: The question directly addresses the topic of the context, making it highly relevant.
        
        Flaws: None.\\

    Complexity: 2/5
    
        Strengths: The question is simple and easy to understand.
        
        Flaws: The simplicity of the question does not challenge the respondent to think deeply about the topic. It could be more complex.\\

    Novelty: 2.5/5\
    
        Strengths: The question is clear and directly related to the topic.
        
        Flaws: The question lacks originality and distinctiveness. It is a basic question that many people might ask when first learning about PPP.\\ 

Strengths in the Question Based on the Evaluation Scores:\\ 

    Grammatical Correctness: The question is well-formed and free from grammatical errors.\\
    
    Relevance: The question directly addresses the topic of the context, making it highly relevant.\\
    
    Appropriateness: The question is semantically correct and understandable.\\

Flaws in the Question Based on the Evaluation Scores:\\ 

    Complexity: The question is quite simple and does not challenge the respondent to think deeply about the topic.\\ 
    
    Novelty: The question is basic and lacks originality. It could be improved by asking something more specific or unique about PPP.\\ 


    
    
    
    

}

\end{tcolorbox}

 \caption{Output provided by $LLM_1$ for a generated question based on Economics from the EduProbe dataset.}
\label{tab:example29}
\end{figure}

\begin{figure}[h]
    \centering
     \begin{tcolorbox}
     [enhanced, fit to height=17cm, 
     colback=blue!5!white, colframe=blue!!white, drop fuzzy shadow]

      {\fontfamily{qcr}\selectfont

      To evaluate the question "What does purchasing power parity do?" based on the human evaluation metrics, we will consider the context and score the question accordingly. Here are the scores and the rationale for each: \\

    Grammaticality: 5/5 
    
        Strengths: The question is grammatically correct and straightforward. 
        
        Flaws: None. \\ 

    Appropriateness: 4.5/5 
    
        Strengths: The question is semantically correct, asking about the function of purchasing power parity (PPP). 
        
        Flaws: The question could be more specific, such as inquiring about particular aspects or impacts of PPP. \\

    Relevance: 5/5 
    
        Strengths: The question is highly relevant to the context, directly asking about purchasing power parity, which is the topic of the given context. 
        
        Flaws: None. \\ 

    Complexity: 2/5 
    
        Strengths: The question is simple and easy to understand. 
        
        Flaws: The simplicity of the question does not require much reasoning or cognitive effort to answer. It could be more complex to encourage deeper thinking. \\ 

    Novelty: 2.5/5 
    
        Strengths: The question is clear and directly related to the topic. 
        
        Flaws: The question lacks originality and distinctiveness, as it is a very basic question that many people might ask when first learning about PPP. \\ 

Strengths in the Question Based on the Evaluation Scores:\\ 

    Grammatical Correctness: The question is well-formed and free from grammatical errors. \\
    
    Relevance: The question directly addresses the topic of the context, making it highly relevant. \\
    
    Appropriateness: The question is semantically correct and understandable. \\ 

Flaws in the Question Based on the Evaluation Scores:\\

    Complexity: The question is quite simple and does not challenge the respondent to think deeply about the topic. \\

    Novelty: The question is basic and lacks originality. It could be improved by asking something more specific or unique about PPP.\\


}

\end{tcolorbox}
\caption{Output provided by $LLM_2$ for a generated question based on Economics from the EduProbe dataset.}
\label{tab:example30}
\end{figure}

\section{Conclusion}
In this work, we propose a novel system, MIRROR (Multi-LLM Iterative Review and Response for Optimized Rating), which utilizes LLMs to automate the evaluation process for open-ended questions generated by AQG systems. Our experiments with state-of-the-art LLMs, including GPT-4, Gemini, and Llama2-70b, show that the scores for the human evaluation metrics namely \textit{relevance}, \textit{appropriateness}, \textit{novelty}, \textit{complexity}, and \textit{grammaticality} improve when using the feedback-based approach called MIRROR and tend to be closer to the human baseline scores. We also observe that Pearson's correlation coefficient between GPT-4 and human experts improves when using MIRROR compared to using direct apparoach for evaluation. Our error analysis shows that \textit{relevance} and \textit{appropriateness} are two metrics where our proposed approach, MIRROR, significantly improves. The results demonstrate that LLMs have the potential to provide a scalable and effective alternative to human evaluation, offering a promising solution for assessing question quality in AQG systems. So far, we have focused on short and medium-sized contexts, whereas future work will focus on applying MIRROR to longer contexts.

\bibliography{references}{}
\bibliographystyle{plainnat}

\newpage

\appendix

\section{Appendix}


\subsection{Results}
Table~\ref{tab:metrics_table_app_1} shows the human evaluation metric scores for the EduProbe dataset. Table~\ref{tab:metrics_t_app_2} shows the human evaluation metric scores for the SciQ dataset. Table~\ref{tab:metrics_t_app_3} shows the Pearson's correlation coefficient scores for the EduProbe and SciQ datasets between GPT-4 and human experts. We observe that the human baseline scores the highest in \textit{grammaticality}, \textit{relevance}, and \textit{appropriateness}, but lower in \textit{novelty} and \textit{complexity} for both the EduProbe and SciQ datasets. GPT-4 performs closest to the human baseline, especially in the feedback-based approach, with high scores in \textit{grammaticality} and \textit{appropriateness}. Gemini consistently scores slightly lower than GPT-4 across all metrics and evaluation methods. Llama2-70b has the lowest scores among the LLMs explored but still performs relatively well, with its highest score in \textit{grammaticality}. Overall, LLMs tend to score higher in the feedback-based approach compared to the direct approach, suggesting that feedback helps improve perceived performance. In summary, GPT-4 outperforms Gemini and Llama2-70b in both evaluation methods, with scores closest to human performance, especially in \textit{grammaticality} and \textit{appropriateness}. Gemini performs moderately well, while Llama2-70b shows relatively lower scores across the metrics. For the EduProbe and SciQ datasets, GPT-4 shows higher correlation scores with human experts when using the feedback-based approach compared to the direct approach across all metrics. The highest correlation score is for \textit{grammaticality} with a feedback-based approach. For the feedback-based approach, \textit{appropriateness} and \textit{relevance} show moderate correlations, suggesting a reasonable alignment with human expert evaluations in these areas. We observe lower correlations in \textit{novelty} and \textit{complexity} because both approaches (i.e., feedback-based and direct) show lower correlations in these areas, with the lowest being \textit{complexity} in the direct approach. The direct approach generally yields lower correlation scores across all metrics, highlighting the importance of feedback in improving GPT-4’s alignment with human expert evaluations. In summary, the correlation scores suggest that GPT-4’s evaluations are more aligned with human experts when using the feedback-based approach, particularly in \textit{grammaticality}. The alignment is moderate in \textit{appropriateness} and \textit{relevance}, while it is weaker in \textit{novelty} and \textit{complexity}. Direct approach results in overall lower correlation scores across all metrics. 

For the SciQ and EduProbe datasets, the feedback-based approach achieves a higher correlation with human evaluations in terms of \textit{grammaticality}, \textit{appropriateness}, \textit{relevance}, \textit{novelty}, and \textit{complexity}, suggesting that iteratively refining the questions using LLM feedback better aligns with human judgments on grammatical correctness. For \textit{appropriateness}, the feedback-based approach shows a modest improvement over the direct approach. This indicates that feedback helps in evaluating whether the questions are suitable and contextually appropriate more closely to how humans would rate them. The feedback-based approach significantly outperforms direct prompting in relevance, suggesting that iterative LLM feedback effectively enhances the model’s ability to judge how relevant the questions are to the given context. The feedback-based method also achieves a higher correlation in assessing \textit{novelty}. This indicates that feedback helps the model better understand and evaluate the uniqueness and originality of the questions. While the improvement in complexity evaluation is smaller, the feedback-based approach still surpasses direct prompting. This suggests that feedback helps the model slightly better assess the intricacy and difficulty of the questions. Overall, the higher correlation scores for the feedback-based approach across all metrics suggest that iteratively using LLM feedback improves the alignment of GPT-4’s evaluations with human expert judgments, making it a more effective method for assessing the quality of generated questions.

\begin{table}
\scriptsize
\centering
\caption{Human evaluation results on the EduProbe dataset for grammaticality (Gram), appropriateness (App), relevance (Rel), novelty (Nov), and complexity (Com). \textcolor{blue}{Blue} indicates the highest metric values for the corresponding methods.}
\begin{tabular}{|l|lllll|}
\hline
\textbf{Model} &  \textbf{Gram} & \textbf{App}  & \textbf{Rel} & \textbf{Nov} & \textbf{Com} 
\\
\hline






Human Baseline & 4.95 & 4.97 & 4.48 & 3.98 & 3.10 \\ \hline



\multicolumn{6}{|c|}{\textit{Eduprobe (Direct Approach)}} \\ \hline

GPT-4  & \textcolor{blue}{4.81} & \textcolor{blue}{4.73} & \textcolor{blue}{4.20} & \textcolor{blue}{4.12} & \textcolor{blue}{3.92} \\ \hline

Gemini  & 4.61 & 4.51 & 4.02 & 4.03 & 3.88  \\ \hline

Llama2-70b & 4.38 & 4.20 & 3.84 & 4.01 & 3.88  \\ \hline

\multicolumn{6}{|c|}{\textit{EduProbe (Feedback-based Approach) }} \\ \hline

GPT-4  & \textcolor{blue}{4.87} & \textcolor{blue}{4.82} & \textcolor{blue}{4.30} & \textcolor{blue}{4.25} & \textcolor{blue}{4.05} \\ \hline

Gemini  & 4.72 & 4.64 & 4.14 & 4.10 & 4.00  \\ \hline

Llama2-70b & 4.60 & 4.62 & 4.08 & 4.06 & 3.83  \\ \hline




















\end{tabular}


\label{tab:metrics_table_app_1}
\end{table}

\begin{table}
\scriptsize
\centering
\caption{Human evaluation results on the SciQ dataset for grammaticality (Gram), appropriateness (App), relevance (Rel), novelty (Nov), and complexity (Com). \textcolor{blue}{Blue} indicates the highest metric values for the corresponding methods.}
\begin{tabular}{|l|lllll|}
\hline
\textbf{Model} &  \textbf{Gram} & \textbf{App}  & \textbf{Rel} & \textbf{Nov} & \textbf{Com} 
\\
\hline






Human Baseline & 4.90 & 4.93 & 4.38 & 3.99 & 3.20 \\ \hline



\multicolumn{6}{|c|}{\textit{SciQ (Direct Approach)}} \\ \hline

GPT-4  & \textcolor{blue}{4.70} & \textcolor{blue}{4.44} & \textcolor{blue}{4.03} & \textcolor{blue}{4.01} & \textcolor{blue}{3.74} \\ \hline

Gemini  & 4.42 & 4.34 & 3.92 & 3.84 & 3.65  \\ \hline

Llama2-70b & 4.23 & 4.10 & 3.73 & 3.67 & 3.28  \\ \hline

\multicolumn{6}{|c|}{\textit{SciQ (Feedback-based Approach) }} \\ \hline

GPT-4  & \textcolor{blue}{4.77} & \textcolor{blue}{4.74} & \textcolor{blue}{4.24} & \textcolor{blue}{4.20} & \textcolor{blue}{4.01} \\ \hline

Gemini  & 4.64 & 4.58 & 4.08 & 4.04 & 3.93  \\ \hline

Llama2-70b & 4.58 & 4.55 & 3.94 & 3.91 & 3.80  \\ \hline




















\end{tabular}


\label{tab:metrics_t_app_2}
\end{table}

\begin{table}
\scriptsize
\centering

\caption{Pearson’s correlation coefficient scores for the EduProbe and SciQ datasets between GPT-4 and human experts on grammaticality (Gram), appropriateness (App), relevance (Rel), novelty (Nov), and complexity (Com). \textcolor{blue}{Blue} denotes the highest correlation values for a particular dataset and approach.}

\begin{tabular}{|l|lllll|}
\hline
\textbf{Model} &  \textbf{Gram} & \textbf{App}  & \textbf{Rel} & \textbf{Nov} & \textbf{Com} 
\\
\hline





\multicolumn{6}{|c|}{\textit{EduProbe}} \\ \hline

GPT-4 (Direct Approach) & 0.41 & 0.32 & 0.26 & 0.25 & 0.28 \\ \hline

GPT-4 (Feedback-based Approach) & \textcolor{blue}{0.62} & \textcolor{blue}{0.48} & \textcolor{blue}{0.51} & \textcolor{blue}{0.42} & \textcolor{blue}{0.38} \\ \hline

\multicolumn{6}{|c|}{\textit{SciQ}} \\ \hline

GPT-4 (Direct Approach) & 0.40 & 0.36 & 0.30 & 0.32 & 0.30 \\ \hline

GPT-4 (Feedback-based Approach) & \textcolor{blue}{0.65} & \textcolor{blue}{0.44} & \textcolor{blue}{0.55} & \textcolor{blue}{0.44} & \textcolor{blue}{0.34} \\ \hline




















\end{tabular}


\label{tab:metrics_t_app_3}
\end{table}

\subsection{Pearson's Correlation Coefficient between LLMs and Human Experts}
\begin{table}[tb]
\scriptsize
\centering

\caption{Pearson's correlation coefficient scores for the EduProbe dataset between LLMs and human experts on \textit{grammaticality} (Gram), \textit{appropriateness} (App), \textit{relevance} (Rel), \textit{novelty} (Nov), and \textit{complexity} (Com). \textcolor{blue}{Blue} denotes the highest correlation values among all approaches and LLMs. \underline{Underline} denotes the better correlation values for a particular LLM among the two approaches (i.e., direct and feedback-based).}

\begin{tabular}{|l|lllll|}
\hline
\textbf{Model} &  \textbf{Gram} & \textbf{App}  & \textbf{Rel} & \textbf{Nov} & \textbf{Com} 
\\
\hline






GPT-4 (Direct Approach) & 0.41 & 0.32 & 0.26 & 0.25 & 0.28 \\ 

GPT-4 (Feedback-based Approach) & \textcolor{blue}{\underline{0.62}} & \textcolor{blue}{\underline{0.48}} & \textcolor{blue}{\underline{0.51}} & \textcolor{blue}{\underline{0.42}} & \textcolor{blue}{\underline{0.38}} \\ \hline

Gemini (Direct Approach) & 0.40 & 0.30 & 0.23 & 0.22 & 0.24 \\ 

Gemini (Feedback-based Approach) & \underline{0.58} & \underline{0.45} & \underline{0.48} & \underline{0.39} & \underline{0.35} \\ \hline

Llama2-70b (Direct Approach) & 0.38 & 0.27 & 0.22 & 0.20 & 0.21 \\ 

Llama2-70b (Feedback-based Approach) & \underline{0.54} & \underline{0.44} & \underline{0.44} & \underline{0.35} & \underline{0.31} \\ \hline




















\end{tabular}


\label{tab:metrics_t_app_11}
\end{table}

\begin{table}[tb]
\scriptsize
\centering
\caption{Pearson's correlation coefficient scores for the SciQ dataset between LLMs and human experts on \textit{grammaticality} (Gram), \textit{appropriateness} (App), \textit{relevance} (Rel), \textit{novelty} (Nov), and \textit{complexity} (Com). \textcolor{blue}{Blue} denotes the highest correlation values among all approaches and LLMs. \underline{Underline} denotes the better correlation values for a particular LLM among the two approaches (i.e., direct and feedback-based).}
\begin{tabular}{|l|lllll|}
\hline
\textbf{Model} &  \textbf{Gram} & \textbf{App}  & \textbf{Rel} & \textbf{Nov} & \textbf{Com} 
\\
\hline






GPT-4 (Direct Approach) & 0.40 & 0.36 & 0.30 & 0.32 & 0.30 \\

GPT-4 (Feedback-based Approach) & \textcolor{blue}{\underline{0.65}} & \textcolor{blue}{\underline{0.44}} & \textcolor{blue}{\underline{0.55}} & \textcolor{blue}{\underline{0.44}} & \textcolor{blue}{\underline{0.34}} \\ \hline

Gemini (Direct Approach) & 0.38 & 0.33 & 0.28 & 0.30 & 0.27 \\

Gemini (Feedback-based Approach) & \underline{0.62} & \underline{0.40} & \underline{0.51} & \underline{0.40} & \underline{0.30} \\ \hline

Llama2-70b (Direct Approach) & 0.36 & 0.30 & 0.25 & 0.27 & 0.24 \\

Llama2-70b (Feedback-based Approach) & \underline{0.60} & \underline{0.37} & \underline{0.48} & \underline{0.36} & \underline{0.28} \\ \hline




















\end{tabular}


\label{tab:metrics_t12}
\end{table}
Table~\ref{tab:metrics_t_app_11} and Table~\ref{tab:metrics_t12} show the results of the Pearson correlation coefficient between LLMs and human experts for the EduProbe and SciQ datasets, respectively. For the EduProbe dataset, all three models show improved correlation scores across all metrics when the feedback-based approach is used compared to the direct approach. GPT-4 shows the most significant improvement in correlation scores with the feedback-based approach, indicating that it benefits the most from receiving feedback. GPT-4 consistently outperforms Gemini and Llama2-70b across all metrics and both approaches (i.e., direct and feedback-based). Gemini performs better than Llama2-70b in all metrics, both with the direct approach and feedback-based approach in terms of correlation scores. GPT-4 with the feedback-based method achieves the highest correlation scores across all metrics, making it the best-performing model in this comparison. The significant increase in scores with the feedback-based approach in all models indicates that providing feedback improves the correlation between human experts and the LLMs.

For the SciQ dataset, starting with GPT-4, the model performs moderately well in terms of correlation scores with the direct approach. However, when the feedback-based approach is applied, there is a noticeable improvement in performance, especially in terms of \textit{grammaticality}, \textit{appropriateness}, and \textit{relevance}. The Gemini model shows a similar trend: its performance under direct prompting is moderate, but it improves with the feedback-based approach. The enhancements are particularly evident in \textit{grammaticality}, \textit{appropriateness}, and \textit{relevance}, though not as pronounced as those seen in GPT-4. Nevertheless, Gemini performs better with the feedback-based approach than with the direct approach in terms of correlation scores. Llama2-70b performs moderately with the direct approach, but feedback-based prompting leads to better correlation scores across all metrics, with the most significant improvements observed in \textit{grammaticality}, \textit{relevance}, and \textit{appropriateness}.


\subsection{Error Analysis}
Table ~\ref{tab:error} shows the percentage of questions with exact matches in scores for different human evaluation metrics using both the direct approach and the feedback-based approach. We conducted a human study based on 100 questions from the EduProbe and SciQ datasets, respectively evaluated by each LLM using both approaches. We observed that the scores provided by human experts matched with the direct approach in 54\%, 39\%, 36\%, 42\%, and 46\% of cases for \textit{grammaticality}, \textit{appropriateness}, \textit{relevance}, \textit{novelty}, and \textit{complexity}, respectively, for GPT-4. In contrast, the scores provided by human experts matched with the feedback-based approach in 67\%, 64\%, 62\%, 55\%, and 61\% of cases for the same metrics. All three LLMs (i.e., GPT-4, Gemini, and Llama2-70b) show a noticeable improvement in their scores when using the feedback-based approach compared to the direct approach, indicating that iterative feedback helps the models perform better. GPT-4 shows the highest scores in both direct and feedback-based methods, making it the most effective model among the three. Gemini and Llama2-70b exhibit similar trends, but Gemini generally scores higher than Llama2-70b in both approaches. For all three LLMs, we observed that \textit{appropriateness} and \textit{relevance} showed the highest improvements with the feedback-based approach compared to the direct approach. These results indicate that providing feedback is particularly beneficial for appropriateness and relevance, significantly enhancing the performance of the LLMs.

\begin{table}[tb]
\scriptsize
\centering
\caption{Percentage of questions showing the exact match in scores for different human evaluation metrics for the direct approach and feedback-based approach. \textcolor{blue}{Blue} denotes the highest percentage values among all approaches and LLMs. \underline{Underline} denotes the better percentage values for a particular LLM among the two approaches (i.e., direct and feedback-based).}
\begin{tabular}{|l|lllll|}
\hline
\textbf{Model} &  \textbf{Gram} & \textbf{App}  & \textbf{Rel} & \textbf{Nov} & \textbf{Com} 
\\
\hline






GPT-4 (Direct Approach) & 54 & 39 & 36 & 42 & 46 \\

GPT-4 (Feedback-based Approach) & \textcolor{blue}{\underline{67}} & \textcolor{blue}{\underline{64}} & \textcolor{blue}{\underline{62}} & \textcolor{blue}{\underline{55}} & \textcolor{blue}{\underline{61}} \\ \hline

Gemini (Direct Approach) & 51 & 36 & 33 & 38 & 41 \\

Gemini (Feedback-based Approach) & \underline{65} & \underline{60} & \underline{57} & \underline{50} & \underline{56} \\ \hline

Llama2-70b (Direct Approach) & 47 & 32 & 30 & 34 & 35 \\

Llama2-70b (Feedback-based Approach) & \underline{62} & \underline{57} & \underline{53} & \underline{45} & \underline{53} \\ \hline




















\end{tabular}


\label{tab:error}
\end{table}

\subsection{Examples of Outputs Generated by LLMs Using the Feedback-based Approach}
The feedback approach consists of feedback occurring between two LLMs, namely $LLM_1$ and $LLM_2$.
Figure ~\ref{tab:example1} shows the output provided by $LLM_1$ on a generated question based on Economics framed from the EduProbe dataset. Figure ~\ref{tab:example2} shows the output provided by $LLM_2$ on a generated question based on Economics framed from the EduProbe dataset. We observe that the scores provided by $LLM_1$ and $LLM_2$ for different human evaluation metrics have become the same after completion of the Feedback Approach.
The context involved in the question generation process is: \textit{"Purchasing power parity (PPP) is an economic indicator that signifies the purchasing power of the currencies of various nations of the world against each other. It helps in comparing living standards between different countries and estimating economic productivity."}. The question generated by GPT-3.5 Turbo is: \textit{"What does purchasing power parity do?"}

Figure ~\ref{tab:example3} shows the output provided by $LLM_1$ on a generated question based on History framed from the EduProbe dataset. Figure ~\ref{tab:example4} shows the output provided by $LLM_2$ on a generated question based on History framed from the EduProbe dataset. We observe that the scores provided by $LLM_1$ and $LLM_2$ for different human evaluation metrics have become same after completion of the Feedback Approach. The context involved in the question generation process is: \textit{"During the medieval period in India, Islamic rulers held significant power, leading to the blending of Indian and Islamic cultures, which can still be observed in the architecture and artwork created at that time. The country was governed and administered by notable rulers who made exceptional contributions in the fields of art, culture, literature, and architecture during this period."} The question generated by GPT-3.5 Turbo is: \textit{"How did medieval period contribute to India?"}

Figure ~\ref{tab:example5} shows the output provided by $LLM_1$ on a generated question based on Biology framed from the SciQ dataset. Figure ~\ref{tab:example6} shows the output provided by $LLM_2$ on a generated question based on Biology framed from the SciQ dataset. We observe that the scores provided by $LLM_1$ and $LLM_2$ for different human evaluation metrics have become same after completion of the Feedback Approach. The context is: "\textit{Mesophiles grow best in moderate temperature, typically between 25'C and 40'C. Mesophiles are often found living in or on the bodies of humans or other animals. The optimal growth temperature of many pathogenic mesophiles is 37".} The question generated by GPT-3.5 Turbo is: "\textit{Where is Mesophiles found?}"

Figure ~\ref{tab:example7} shows the output provided by $LLM_1$ on a generated question based on Earth Science framed from the SciQ dataset. Figure ~\ref{tab:example8} shows the output provided by $LLM_2$ on a generated question based on Earth science framed from the SciQ dataset. We observe that the scores provided by $LLM_1$ and $LLM_2$ for different human evaluation metrics have become same after completion of the Feedback Approach. The context is: "\textit{Without Coriolis Effect the global winds would blow north to south or south to north. But Coriolis
makes them blow north-east to southwest or the reverse in the Northern Hemisphere. The winds blow northwest to southeast or the reverse in the southern hemisphere.}"
The question generated by GPT-3.5 Turbo is: "\textit{How does Coriolis effect impact global winds?}"

\begin{figure}[h]
    \centering
     \begin{tcolorbox}
     [enhanced, fit to height=19.5cm, 
     colback=teal!5!white, colframe=teal!!white, drop fuzzy shadow]

      {\fontfamily{qcr}\selectfont

      Scores and Rationale for the Question: "What does purchasing power parity do?" \\ 

    Grammaticality: 5/5
    
        Strengths: The question is well-formed and free from grammatical errors.

        Flaws: None.\\

    Appropriateness: 4.5/5
    
        Strengths: The question is semantically correct and understandable.
        
        Flaws: The question is very general and could be more specific about PPP.\\

    Relevance: 5/5
    
        Strengths: The question directly addresses the topic of the context, making it highly relevant.
        
        Flaws: None.\\

    Complexity: 2/5
    
        Strengths: The question is simple and easy to understand.
        
        Flaws: The simplicity of the question does not challenge the respondent to think deeply about the topic. It could be more complex.\\

    Novelty: 2.5/5\
    
        Strengths: The question is clear and directly related to the topic.
        
        Flaws: The question lacks originality and distinctiveness. It is a basic question that many people might ask when first learning about PPP.\\ 

Strengths in the Question Based on the Evaluation Scores:\\ 

    Grammatical Correctness: The question is well-formed and free from grammatical errors.\\
    
    Relevance: The question directly addresses the topic of the context, making it highly relevant.\\
    
    Appropriateness: The question is semantically correct and understandable.\\

Flaws in the Question Based on the Evaluation Scores:\\ 

    Complexity: The question is quite simple and does not challenge the respondent to think deeply about the topic.\\ 
    
    Novelty: The question is basic and lacks originality. It could be improved by asking something more specific or unique about PPP.\\ 


    
    
    
    

}

\end{tcolorbox}

 \caption{Output provided by $LLM_1$ for a generated question based on Economics from the EduProbe dataset.}
\label{tab:example1}
\end{figure}

\begin{figure}[h]
    \centering
     \begin{tcolorbox}
     [enhanced, fit to height=20cm, 
     colback=blue!5!white, colframe=blue!!white, drop fuzzy shadow]

      {\fontfamily{qcr}\selectfont

      To evaluate the question "What does purchasing power parity do?" based on the human evaluation metrics, we will consider the context and score the question accordingly. Here are the scores and the rationale for each: \\

    Grammaticality: 5/5 
    
        Strengths: The question is grammatically correct and straightforward. 
        
        Flaws: None. \\ 

    Appropriateness: 4.5/5 
    
        Strengths: The question is semantically correct, asking about the function of purchasing power parity (PPP). 
        
        Flaws: The question could be more specific, such as inquiring about particular aspects or impacts of PPP. \\

    Relevance: 5/5 
    
        Strengths: The question is highly relevant to the context, directly asking about purchasing power parity, which is the topic of the given context. 
        
        Flaws: None. \\ 

    Complexity: 2/5 
    
        Strengths: The question is simple and easy to understand. 
        
        Flaws: The simplicity of the question does not require much reasoning or cognitive effort to answer. It could be more complex to encourage deeper thinking. \\ 

    Novelty: 2.5/5 
    
        Strengths: The question is clear and directly related to the topic. 
        
        Flaws: The question lacks originality and distinctiveness, as it is a very basic question that many people might ask when first learning about PPP. \\ 

Strengths in the Question Based on the Evaluation Scores:\\ 

    Grammatical Correctness: The question is well-formed and free from grammatical errors. \\
    
    Relevance: The question directly addresses the topic of the context, making it highly relevant. \\
    
    Appropriateness: The question is semantically correct and understandable. \\ 

Flaws in the Question Based on the Evaluation Scores:\\

    Complexity: The question is quite simple and does not challenge the respondent to think deeply about the topic. \\

    Novelty: The question is basic and lacks originality. It could be improved by asking something more specific or unique about PPP.\\


}

\end{tcolorbox}
\caption{Output provided by $LLM_2$ for a generated question based on Economics from the EduProbe dataset.}
\label{tab:example2}
\end{figure}

\begin{figure}[h]
    \centering
     \begin{tcolorbox}
     [enhanced, fit to height=22cm, 
     colback=teal!5!white, colframe=teal!!white, drop fuzzy shadow]

      {\fontfamily{qcr}\selectfont

      To evaluate the question "How did medieval period contribute to India?" based on the human evaluation metrics, we will consider the context and score the question accordingly. Here are the scores and the rationale for each:\\ 

    Grammaticality: 3.5/5 
    
        Strengths: The question is mostly grammatically correct.
        
        Flaws: The article "the" is missing before "medieval period," which makes the question slightly awkward. It should read: "How did the medieval period contribute to India?"\\ 

    Appropriateness: 4.5/5
    
        Strengths: The question is semantically correct, asking about the contributions of the medieval period. 
        
        Flaws: The wording could be more precise, such as specifying which aspects of India (e.g., culture, art, architecture). \\ 

    Relevance: 5/5 
    
        Strengths: The question is highly relevant to the context, directly inquiring about the contributions of the medieval period, which includes significant cultural, artistic, and administrative aspects. 
        
        Flaws: None. \\ 

    Complexity: 4/5 
    
        Strengths: The question requires a detailed and thoughtful response, considering various contributions in multiple fields (art, culture, literature, architecture).
        
        Flaws: The question could be more complex by asking about specific contributions or comparing them to those of other periods.\\ 

    Novelty: 3.5/5 
    
        Strengths: The question is clear and prompts a discussion on an important historical period.
        
        Flaws: While the question is pertinent, it is somewhat general and could be seen as a common inquiry in the study of history.\\ 

Strengths in the Question Based on the Evaluation Scores:\\ 

    Relevance: The question directly addresses the topic of the context, making it highly relevant.\\ 
    
    Appropriateness: The question is semantically correct and understandable.\\ 
    
    Complexity: The question prompts a detailed response, requiring consideration of multiple aspects of the medieval period.\\

Flaws in the Question Based on the Evaluation Scores:\\

    Grammaticality: The question is slightly awkward due to the missing article "the" before "medieval period." \\ 
    
    Novelty: The question, while clear and pertinent, is somewhat general and lacks a unique angle. It could be improved by focusing on specific contributions or comparing the medieval period's contributions to those of other periods. \\ 
    
    Complexity: Although reasonably complex, it could be further detailed to enhance the level of reasoning or cognitive effort required to answer.\\ 

}

\end{tcolorbox}
\caption{Output provided by $LLM_1$ for a generated question based on History from the EduProbe dataset.}
\label{tab:example3}
\end{figure}

\begin{figure}[h]
    \centering
     \begin{tcolorbox}
     [enhanced, fit to height=22cm, 
     colback=blue!5!white, colframe=blue!!white, drop fuzzy shadow]

      {\fontfamily{qcr}\selectfont


To evaluate the question "How did medieval period contribute to India?" based on the human evaluation metrics, we will consider the context and score the question accordingly. Here are the scores and rationale for each metric:\\

    Grammaticality: 3.5/5
    
        Strengths: The question is mostly grammatically correct.
        
        Flaws: The article "the" is missing before "medieval period," making the question slightly awkward. It should read: "How did the medieval period contribute to India?"\\ 

    Appropriateness: 4.5/5
    
        Strengths: The question is semantically correct, asking about the contributions of the medieval period.
        
        Flaws: The wording could be more precise, such as specifying which aspects of India (e.g., culture, art, architecture).\\ 

    Relevance: 5/5
    
        Strengths: The question is highly relevant to the context, directly inquiring about the contributions of the medieval period, which includes significant cultural, artistic, and administrative aspects. 
        
        Flaws: None.\\ 

    Complexity: 4/5 
    
        Strengths: The question requires a detailed and thoughtful response, considering various contributions in multiple fields (art, culture, literature, architecture).  
        
        Flaws: The question could be more complex by asking about specific contributions or comparing them to those of other periods. \\ 

    Novelty: 3.5/5 
    
        Strengths: The question is clear and prompts a discussion on an important historical period. 
        
        Flaws: While the question is pertinent, it is somewhat general and could be seen as a common inquiry in the study of history. \\ 

Strengths in the Question Based on the Evaluation Scores:\\

Relevance: The question is highly pertinent to the context, directly addressing the main topic.\\ 

Appropriateness: The question is clear and semantically accurate.\\ 

Complexity: The question encourages a detailed response, requiring consideration of various aspects of the medieval period. \\

Flaws in the Question Based on the Evaluation Scores: \\ 

Grammaticality: The question is slightly awkward due to the missing article "the" before "medieval period."\\ 

Novelty: While clear and relevant, the question is somewhat general and lacks a unique angle. It could be improved by focusing on specific contributions or comparing the medieval period's contributions to those of other periods.\\ 

Complexity: Although reasonably complex, the question could be more detailed to enhance the level of reasoning or cognitive effort required to answer.\\ 

}

\end{tcolorbox}
\caption{Output provided by $LLM_2$ for a generated question based on History from the EduProbe dataset.}
\label{tab:example4}
\end{figure}

\begin{figure}[h]
    \centering
     \begin{tcolorbox}
     [enhanced, fit to height=11cm, 
     colback=teal!5!white, colframe=teal!!white, drop fuzzy shadow]

      {\fontfamily{qcr}\selectfont
Evaluation Scores:\\

    Grammaticality: 5/5 
    
    Appropriateness: 5/5
    
    Relevance: 5/5
    
    Complexity: 3/5
    
    Novelty: 4/5\\

Strengths in the question based on the evaluation scores provided above:\\

    Grammaticality: The question is grammatically correct, with proper sentence structure and punctuation.\\ 
    
    Appropriateness: The question is semantically correct, making sense in the context of the given information.\\ 
    
    Relevance: The question is highly relevant to the context provided, directly addressing the key point about the optimal growth temperature of mesophiles.\\ 
    
    Novelty: The question is somewhat original and distinct, asking both for the optimal growth temperature and the reason for its significance, which adds depth to the inquiry.\\ 

Flaws in the question based on the evaluation scores provided above:\\ 

    Complexity: The question is relatively straightforward and does not require a high level of reasoning or cognitive effort to answer. A more complex question might delve deeper into the implications of the optimal growth temperature or its effects on human health.\\

}

\end{tcolorbox}
\caption{Output provided by $LLM_1$ for a generated question based on Biology from the SciQ dataset.}
\label{tab:example5}
\end{figure}

\begin{figure}[h]
    \centering
     \begin{tcolorbox}
     [enhanced, fit to height=11cm, 
     colback=blue!5!white, colframe=blue!!white, drop fuzzy shadow]

      {\fontfamily{qcr}\selectfont
Evaluation Scores:\\ 

    Grammaticality: 5/5
    
    Appropriateness: 5/5
    
    Relevance: 5/5
    
    Complexity: 3/5 
    
    Novelty: 4/5 \\

Strengths in the question based on the evaluation scores provided above:\\ 

    Grammaticality: The question is grammatically accurate, with correct sentence structure and punctuation.\\ 
    
    Appropriateness: The question is semantically accurate and fits well with the given context.\\ 
    
    Relevance: The question is highly relevant to the provided context, directly addressing the key point about the optimal growth temperature of mesophiles.\\

    Novelty: The question is somewhat original and unique, as it asks not only about the optimal growth temperature but also why it is significant, adding depth to the inquiry.\\ 

Flaws in the question based on the evaluation scores provided above:\\ \\

    Complexity: The question is relatively simple and does not require a high level of reasoning or cognitive effort to answer. A more complex question could explore the implications of the optimal growth temperature or its effects on human health.\\

}

\end{tcolorbox}
\caption{Output provided by $LLM_2$ for a generated question based on Biology from the SciQ dataset.}
\label{tab:example6}
\end{figure}

\begin{figure}[h]
    \centering
     \begin{tcolorbox}
     [enhanced, fit to height=11cm, 
     colback=teal!5!white, colframe=teal!!white, drop fuzzy shadow]

      {\fontfamily{qcr}\selectfont
Evaluation Scores:\\ 

    Grammaticality: 5/5
    
    Appropriateness: 5/5
    
    Relevance: 4/5
    
    Complexity: 2/5
    
    Novelty: 2/5\\

Strengths in the question based on the evaluation scores provided above:\\ 

    Grammaticality: The question is free of grammatical errors, with proper sentence structure and punctuation.\\ 
    
    Appropriateness: The question is semantically accurate and fits the context.\\ 
    
    Relevance: The question is pertinent to the context, as it asks about the Coriolis effect mentioned in the provided information.\\

Flaws in the question based on the evaluation scores provided above:\\ 

    Complexity: The question is very straightforward and does not require significant reasoning or cognitive effort to answer. It could be more challenging by asking for a detailed explanation or the implications of the Coriolis effect.\\ 
    
    Novelty: The question lacks originality and distinctiveness, as it is a basic inquiry that doesn't explore unique aspects or implications of the Coriolis effect mentioned in the context.\\

}

\end{tcolorbox}
\caption{Output provided by $LLM_1$ for a generated question based on Earth Science from the SciQ dataset.}
\label{tab:example7}
\end{figure}

\begin{figure}[h]
    \centering
     \begin{tcolorbox}
     [enhanced, fit to height=11cm, 
     colback=blue!5!white, colframe=blue!!white, drop fuzzy shadow]

      {\fontfamily{qcr}\selectfont

Evaluation Scores:\\

    Grammaticality: 5/5
    
    Appropriateness: 5/5
    
    Relevance: 4/5
    
    Complexity: 2/5
    
    Novelty: 2/5\\

Strengths in the question based on the evaluation scores provided above:\\

    Grammaticality: The question is grammatically correct, with no errors in sentence structure or punctuation.\\ 
    
    Appropriateness: The question is semantically correct and makes sense given the context. \\ 
    
    Relevance: The question is relevant to the context, as it directly asks about the Coriolis effect mentioned in the provided information.\\ 

Flaws in the question based on the evaluation scores provided above:\\

    Complexity: The question is very simple and does not require much reasoning or cognitive effort to answer. It could be improved by asking for more detailed information or explanation.\\ 
    
    Novelty: The question lacks originality and distinctiveness, as it is a very basic inquiry that does not delve into the unique aspects or implications of the Coriolis effect described in the context.\\

}

\end{tcolorbox}
\caption{Output provided by $LLM_2$ for a generated question based on Earth Science from the SciQ dataset.}
\label{tab:example8}
\end{figure}

\subsection{Implementation Details}
\label{implementation_details}
For GPT-3.5 Turbo we use a \texttt{temperature} value of 0.7 and \texttt{max\_tokens} size of 2048. For GPT-4 we use a \texttt{temperature} value of 0.7 and \texttt{max\_tokens} size of 2048. For Llama2-70b we use a \texttt{temperature} value of 0.8 and \texttt{max\_tokens} size of 2048. For Gemini we use a \texttt{temperature} value of 0.7 and \texttt{max\_tokens} size of 2048.


\end{document}